# A Deep Learning Approach for Forecasting Air Pollution in South Korea Using LSTM

**Tien-Cuong Bui**      **Van-Duc Le**      **Sang K. Cha**

Department of Electrical and Computer Engineering
Seoul National University
{cuongbt91, levanduc, chask}@snu.ac.kr

**Abstract.** Over the last few years, tackling air pollution is an urgent problem in South Korea. Much research is being conducted in environmental science to evaluate the severe impact of particulate matters on public health. Besides that, deterministic models of air pollutant behavior are also generated; however, these are both complex and often inaccurate. On the contrary, deep recurrent neural network reveals strong potential on forecasting outcomes of time-series data and has become more prevalent. This paper uses Recurrent Neural Networks and Long Short-Term Memory units as a framework for leveraging knowledge from time-series data of air quality and meteorological information. Finally, we investigate prediction accuracies of various configurations. This paper is a significant motivation for not only continuing researching on urban air quality but also helping the government leverage that insight to enact beneficial policies.

## 1. Introduction

Air pollution has been the subject of many present environmental studies due to the inclination of industrialization as well as a temporal correlation between long-term exposure to fine particulate matter and acute increases in mortality including lung cancer and cardiopulmonary [8]. Recently, South Korea has joined the rank of the most polluted countries, and Seoul associated with Beijing and Delhi are the world three most polluted in daily rankings [2]. Additionally, the increment of urbanization plays a vital role in exposing public health to this lethal problem. The primary air pollutants in urban areas include carbon dioxide ($CO_2$), carbon monoxide (CO), nitrogen oxides ($NO_2$), nitrogen monoxide (NO), and particulate matters $PM_{2.5}$, $PM_{10}$. However, the most concerned air pollution factor is $PM_{2.5}$ or particulate matter that is up to 2.5 microns in diameter. These particles are tiny and light allowing them to stay in the atmosphere for a more extended period and also easily bypass the filters of human nose and throat due to their size property. According to C. Arden Pope [8], each 10-g/m3 elevation in long-term average $PM_{2.5}$ ambient concentrations was associated with approximately 4-8 percent increased the risk of cardiopulmonary and lung cancer mortality.

Many researchers from environmental to data science have proposed various solutions for forecasting air pollution using either statistical or deep learning models. Our research uses Encoder - Decoder model that is prominent in Natural Language

Processing problems. Besides that, since the collected data are represented as time series or hourly timesteps, we apply Recurrent Neural Networks with Long Short-Term Memory units to the prediction model. However, forecasting $PM_{2.5}$ AQI values is not an easy problem since air pollution may be impacted by many factors and it does not firmly relate to the past repetitive patterns. The main contributions of this paper are (1) Provide two datasets comprised of meteorological and air quality information in South Korea (2) Propose a deep learning approach for forecasting air quality index.

## 2. Methods and Models

In this section, we will provide detailed information to the Encoder-Decoder (En-De) model [7] and Long Short-Term Memory [6] units used in our research.

### 2.1. Long-Short Term Memory Units

Given an input sequence $x = (x_1, \ldots, x_T)$, a standard recurrent neural network (RNN) computes the hidden vector sequence $h = (h_1, \ldots, h_T)$ and output vector sequence $y = (y_1, \ldots, y_T)$ by iterating the following equations from t = 1 to T:

$$h_t = \mathcal{H}(W_{xh}x_t + W_{hh}h_{t-1} + b_h) \quad (1)$$
$$y_t = W_{hy}h_t + b_y \quad (2)$$

where the $W$ terms denote weight matrices (e.g., $W_{xh}$ is the input-hidden weight matrix), the $b$ terms denote bias vectors (e.g., $b_h$ is hidden bias vector) and $\mathcal{H}$ is a hidden layer function. $\mathcal{H}$ is usually an element-wise application of a sigmoid function. A conventional RNN often confronts poor correctness when executing long sequences due to the explosion gradient problem. Replacing a powerless logistic function with Long Short-Term Memory (LSTM) units can solve this issue. Although many LSTM architectures differ in their structure and activation functions, all of them have explicit memory cells with complicated dynamics allowing it to easily "memorize" information for an extended number of timesteps. The LSTM architecture used in our experiments is given by the following equations:

$$i_t = \sigma(W_{xi}x_t + W_{hi}h_{t-1} + b_i) \quad (3)$$
$$f_t = \sigma(W_{xf}x_t + W_{hf}h_{t-1} + b_f) \quad (4)$$
$$o_t = \sigma(W_{xo}x_t + W_{ho}h_{t-1} + b_o) \quad (5)$$
$$c_t = f_t * c_{t-1} + i_t * \tanh(W_{xg}x_t + b_g) \quad (6)$$
$$h_t = o_t * \tanh(c_t) \quad (7)$$

where $\sigma$ is the logistic sigmoid function, and $i, f, o$, and $c$ are respectively the input gate, forget gate, output gate, and cell activation vectors, all of which are the same size as the hidden vector $h$. The weight matrices from the cell to gate vectors (e.g., $W_{xi}$) are diagonal, so element $m$ in each gate vector only receives input from element $m$ of the cell vector.

## 2.2. Encoder-Decoder model

In the model, an encoder reads the sequence of vectors $x = (x_1, ..., x_T)$ into a vector $c_i$ using RNN with LSTM units as mentioned in section 2.1.

$$h_t = f(x_t; h_{t-1}) \qquad (8)$$
$$c_i = q(\{h_1, ..., h_T\}) \qquad (9)$$

where $h_t \in R^n$ is a hidden state at time t, and $c_i$ is a vector generated from a sequence of hidden vectors, and $f$ & $q$ are some nonlinear functions. This model uses an RNN with LSTM units as $f$ and a mean operation as $q$.

A decoder is often trained to predict the value $y_{t'}$ given the context vector $c_i$ and all the previous predicted values $\{y_1, ..., y_{t'-1}\}$. In other words, a conventional decoder defines a probability over the prediction $y$ by decomposing the joint probability into the conditional order:

$$p(\mathbf{y}) = \prod_{t=1}^{T} p(y_t \mid \{y_1, ..., y_{t-1}\}, c_i) \qquad (10)$$

where $\mathbf{y} = (y_1, ..., y_{T_y})$. With an RNN, each conditional probability is modeled as follows:

$$p(y_t \mid \{y_1, ..., y_{t-1}\}, c) = g(y_{t-1}, s_t, c_i) \qquad (11)$$

where $g$ is a nonlinear, potentially multi-layered, function that outputs the probability of $y_t$, and $s_t$ is the hidden state of the RNN. Encoder-Decoder model can stack multiple RNN layers on the top of each other to increase correctness.

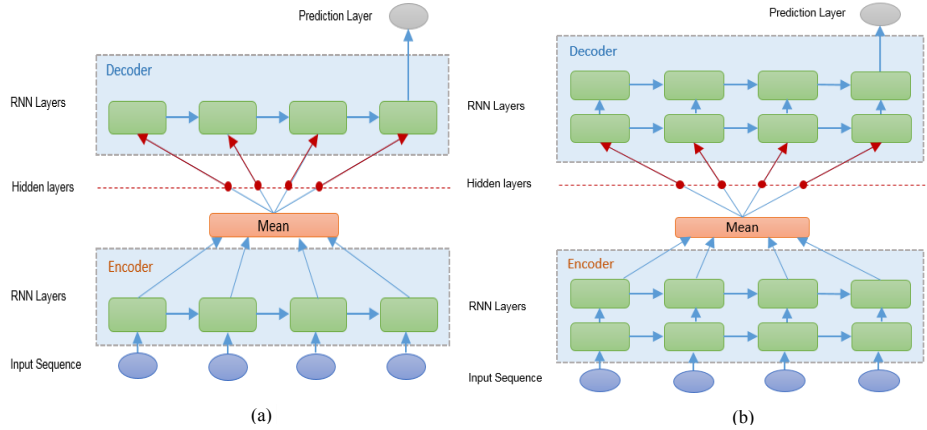

(a)      (b)

**Fig. 1.** Encoder-Decoder networks, (a) has a single RNN layer whereas (b) has two RNN layers stacked on the top of each other

For training, ADAM [5] algorithm is used as the gradient-based optimizer, instead of SGD. ADAM outperforms SGD regarding faster convergence and lower error ratios.

## 3. Dataset and Experiments

The presented method was implemented on two datasets collected from multiple resources. More specially, weather data from July 2008 to April 2018 of researched cities were crawled from World Weather Online and Time & Date while AQI information was collected from Air Korea, Seoul Clean Air, Chinese PM$_{2.5}$ report, and Daegu Environment Government private APIs. Next, Spark was used to clean irrelevant data, fulfill missing elements, join data from various resources to one jointly dataset, and transform context features into vectors, in which, each vector corresponds to one-hour timestep. In particular, a rough dataset of Daegu is approximately 10Gb with more than 30 million records. After pre-processing, the analyzing engine kept only 27.936 rows. The dataset was separated into two parts, in which, the first part related to the period from January 2015 to June 2017 due to the missing of air quality information, and the second one was from June 2017 to March 2018. At first, we trained the model on the first part then transfers the learned weights to the second one instead of training from scratch. The period from February to March 10th, 2018 was used as the testing set. On the contrary, Seoul data included complete information from January 2008 to April 2018 with more than 2 million records corresponding to hourly records of 25 districts. Additionally, on Seoul dataset, the model did not use transfer method; data from 2008 to 2016 was used as training set, and test set was the data from January 2017 to April 2018. Both datasets were randomly carved out twenty percent of the training set to be the validation set.

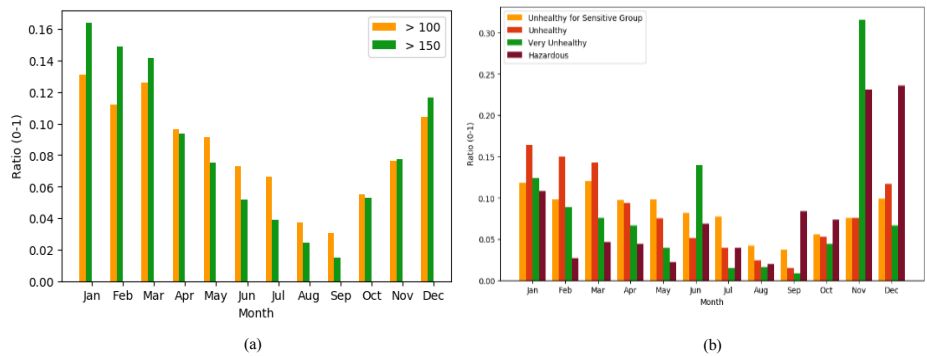

**Fig. 2.** Shows Monthly Distribution of PM$_{2.5}$ AQI in Seoul (2008-2018). (a) Shows only the aggregation of the Unhealthy levels when PM$_{2.5}$ AQI is above 100 and 150, (b) reveals all four warning level distribution. As shown in Figure 3, dangerous categories increase substantially from October of a current year to March of the next year and reach the peak point in January. After that, it gradually declines from March to the lowest point in September of the next year. Consequently, months in a year, hours in a day, holidays information were also elements of featured vectors.

According to the collected datasets, we found that in a short period ($\leq 5$ hours), PM$_{2.5}$ AQI values fluctuate slightly due to the similarity in meteorological conditions. Therefore, the prediction machine can adjust accurately to the future outcome causing the results to be useless. Consequently, we only focused on forecasting long-term state of air pollution ($\geq 8$ hours).

**Table 1.** RMSE values of different settings

| Model settings | 8h | 12h | 16h | 20h | 24h |
|---|---|---|---|---|---|
| TF + RNN + MAE | **12.41** | 13.48 | 14.48 | 14.42 | 17.30 |
| TF + RNNs + MAE | 13.47 | **13.19** | **12.85** | **12.79** | **13.54** |
| Joint + RNN + MAE | 13.84 | 13.18 | 13.19 | 13.16 | 13.79 |
| Joint + RNNs + MAE | 15.17 | 14.82 | 14.71 | 14.85 | 15.12 |
| TF + RNN + MSE | 13.96 | 13.51 | 13.47 | 13.43 | 14.02 |
| Joint + RNN + MSE | 13.73 | 13.24 | 13.10 | 13.30 | 14.47 |
| TF + RNNs + MSE | 14.26 | 14.34 | 14.02 | 13.89 | 15.20 |
| Joint + RNNs + MSE | 15.99 | 15.18 | 14.76 | 14.58 | 14.85 |

Table 1 reveals that using MAE in our training model is more effective than MSE, and transfer method is appropriate for Daegu dataset. Consequently, the combination of multiple RNN layers model, MAE loss function, and transfer method provides outstanding prediction results.

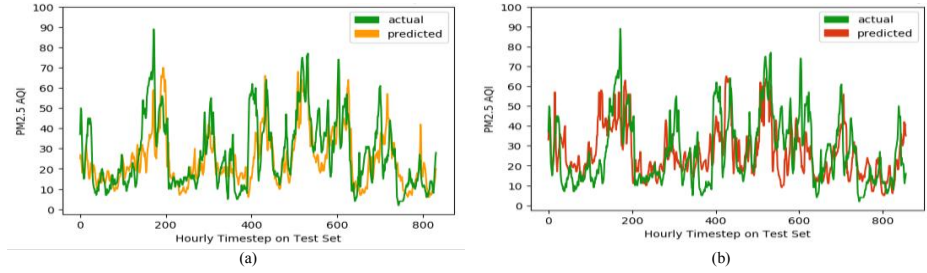

**Fig. 3.** (a) shows the predictions of a single RNN model, (b) presents predictions of two RNN layers model; both use 24 encoding timesteps, 8 decoding timesteps, MAE, and transfer method. As shown in Table 1, the RMSE values are 12.41 & 13.47

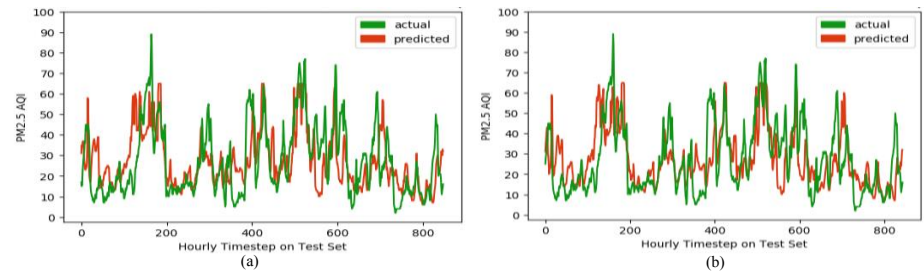

**Fig. 4.** Shows prediction results of two RNN layers En-De 24h past to 16 & 20 future timesteps using transfer method and MAE loss function on training. From Table 1, the RMSE values are 12.85, 12.79

Next, we measured the accuracy of the proposed model on Seoul Dataset, in which we conducted three experiments. Firstly, the prediction machine learned and generated predictions on a complete dataset including full features, and used two models similar to Daegu configurations. After that, the model was re-trained on a dataset, which was removed all China-related features. In the latter experiment, only two RNN layers model was used.

Table 2 presents that a single RNN model performs efficiently only on forecasting 8 hours. When detaching all China-related features from the dataset, the prediction accuracies significantly decline, even though the training model uses two RNN layers model. In conclusion, utilizing a robust model and joining complete China-related features to training vectors provides highest accuracies.

**Table 2:** Test RMSE for 2017-2018 with different settings

| Model settings | 8h | 12h | 16h | 20h | 24h |
| --- | --- | --- | --- | --- | --- |
| Joint Dataset + RNNs | 26.27 | **27.08** | **28.61** | **29.97** | **31.29** |
| Joint Dataset + RNN | **25.93** | 27.61 | 29.05 | 30.21 | 31.52 |
| Seoul Dataset+ RNNs | 27.51 | 28.49 | 29.66 | 30.6 | 31.8 |

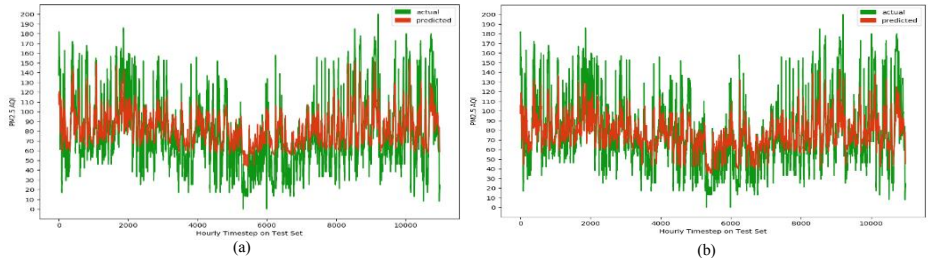

(a)  (b)

**Fig. 5.** Shows the predicted values of future 8 & 12 hours (a) using a single RNN, (b) using two RNN layers. As shown in Table 2, the RMSE scores are **25.93** and **27.08**

## 4. Conclusion

The goal of the presented work was to evaluate the effectiveness of encoder-decoder networks for building prediction machines with time series data. The proposed model shows significant results in prediction $PM_{2.5}$ AQI of long future based on historical meteorological data. However, to enhance the accuracy of the prediction machine, the model needs to be evaluated more in the future. Finally, forecasting the status of air pollution can help governments in policy-making and resource allocation.